\documentclass[letterpaper, 10 pt, conference]{ieeeconf}
\IEEEoverridecommandlockouts                          
\usepackage{amsmath,amssymb,amsfonts}
\usepackage{algorithmic}
\usepackage{graphicx}
\usepackage{textcomp}
\usepackage{xcolor}
\usepackage{bm}
\usepackage{tikz}

\title{\LARGE \bf
Learning-Based Design of Off-Policy Gaussian Controllers: \\ Integrating Model Predictive Control and Gaussian Process Regression
}

\author{Shiva Kumar Tekumatla, Varun Gampa, and Siavash Farzan%
\thanks{Shiva Kumar Tekumatla and Varun Gampa are with the Robotics Engineering Department, Worcester Polytechnic Institute, Worcester, MA 01609, USA, {\tt\small \{stekumatla,vgampa\}@wpi.edu}}%
\thanks{Siavash Farzan is with the Electrical Engineering Department, California Polytechnic State University, San Luis Obispo, CA 93407, USA, {\tt\small sfarzan@calpoly.edu}}
}

\begin{document}

\maketitle
\thispagestyle{empty}
\pagestyle{empty}

\begin{abstract}
This paper presents an off-policy Gaussian Predictive Control (GPC) framework aimed at solving optimal control problems with a smaller computational footprint, thereby facilitating real-time applicability while ensuring critical safety considerations. The proposed controller imitates classical control methodologies by modeling the optimization process through a Gaussian process and employs Gaussian Process Regression to learn from the Model Predictive Control (MPC) algorithm. Notably, the Gaussian Process setup does not incorporate a built-in model, enhancing its applicability to a broad range of control problems. We applied this framework experimentally to a differential drive mobile robot, tasking it with trajectory tracking and obstacle avoidance. Leveraging the off-policy aspect, the controller demonstrated adaptability to diverse trajectories and obstacle behaviors. Simulation experiments confirmed the effectiveness of the proposed GPC method, emphasizing its ability to learn the dynamics of optimal control strategies. Consequently, our findings highlight the significant potential of off-policy Gaussian Predictive Control in achieving real-time optimal control for handling of robotic systems in safety-critical scenarios.
\end{abstract}

\section{Introduction}

In recent years, there has been a marked increase in the utilization of learning-based control methods due to their ability to learn the complex nature of classical control methods~\cite{brunke2021safe}, coupled with their flexibility and adaptability, all while typically demanding lower computing power. Most classical control methods require solving an optimal control problem subject to underlying constraints. For instance, in Model Predictive Control (MPC)~\cite{MPHC}, the optimal control problem must be solved repeatedly at each state update to derive a control policy that maximizes reward or minimizes costs. An important question arising in this context is whether it is plausible to approximate the optimal control problem to a function. More explicitly, is it possible to approximate the numerical optimization intrinsic to a control problem?

In this paper, we propose a novel approach designed to replace the MPC optimization process, offering a balance of computational efficiency and efficacy. This methodology leverages learning-based techniques to gain insights into the optimization dynamics inherent in optimal control strategies, such as MPCs. Specifically, we model the numerical optimization process as a Gaussian process and formulate an off-policy Gaussian Predictive Controller (Off-Policy GPC) that learns from the MPC algorithm's behavior on a given robotic platform. Remarkably, the proposed controller is designed not to incorporate system dynamics, learning solely from the behavior of the MPC algorithm, making it applicable to diverse robot platforms.

The controller is adept at tracking prescribed trajectories and avoiding obstacles, and its capacity to learn from a diverse set of trajectories allows it to learn about the robot dynamics and how to interact with the environment.
We present the theoretical foundations for formulating the off-policy Gaussian process control derived from MPC, explaining how it enables the learning of the overall dynamics of both the environment and the robot through the controller. Practical implementation aspects of motion planning and control are also discussed, along with the comparative benefits of our proposed GPC approach over traditional MPC methods.

Our contributions include proposing a generalized off-policy learning algorithm, based on Gaussian Processes, that combines optimal control methods with learning capabilities to understand the behavior of MPC. Instead of iteratively solving optimal control problems, the proposed architecture leverages existing data for inferential needs as they arise. While the robot model is integral to the MPC, it is extraneous to the learning architecture. To evaluate the effectiveness of our approach, we conduct extensive simulations utilizing a differential drive mobile robot. The results illustrate that the off-policy Gaussian Predictive Control either mirrors or surpasses MPC performance in terms of tracking accuracy and obstacle avoidance, underscoring the potential of the proposed methodology to achieve optimal control in complex robotic systems with enhanced real-time performance.

This paper is structured as follows: Section~\ref{sec:related-work} explores existing work in the field and outlines the inspiration behind our approach. Section \ref{sec:method} formulates the off-policy GPC architecture and illustrates its reliance on MPC and Gaussian processes, which serve as the foundations upon which the framework is built. Section \ref{sec:env} details the experimental setup and the dynamics of the robot. Section \ref{sec:results} presents and discussed the simulation results, and finally, Section \ref{sec:conclusions} concludes the paper, highlighting challenges and avenues for future work.

\section{Related Work} \label{sec:related-work}
Optimal control problems in real-world applications predominantly encounter the stochastic nature of the environment. Consequently, any mathematical tool purposed to approximate the behavior of control methods must adequately accommodate this stochastic behavior. Gaussian Processes (GPs) are notably proficient in approximating models influenced by stochastic data~\cite{ebden2015gaussian}. Although the training of Gaussian Process regression is nonparametric and can be more time-intensive than its linear regression counterpart, it outperforms in capturing the comprehensive behavior of the model~\cite{Melo2012GaussianPF}. The probabilistic non-parametric approach of Gaussian process models is instrumental in highlighting regions of low prediction quality, attributed to either insufficient data or inherent complexity.

Much of the existing literature involving GPs in the context of robotics primarily focuses on learning uncertainty within dynamics or enhancing classical control methods. Kocijan et al.~\cite{GPMPC} discuss the application of Gaussian process models for identifying non-linear dynamic systems, expounding the advantages inherent in model-based predictive control. This research accentuates the pivotal role of variance information derived from Gaussian process models in optimizing control signals.
The study presented in~\cite{deisenroth2011pilco} unfolds a method named Pilco, a practical and data-efficient model-based policy search method. It addresses the significant challenge of model bias inherent in model-based reinforcement learning. By utilizing GPs to learn a probabilistic dynamics model, Pilco adeptly incorporates model uncertainty into long-term planning, allowing for learning from minimal data and fostering proficiency from scratch within a limited number of trials. Drawing inspiration from Pilco, Yoo et al.~\cite{hybrid-pilco} engineered a hybrid controller. This innovative design combines reinforcement learning with deterministic controllers, exemplified in the learning of quadrotor control. It integrates a classical controller, such as PD or LQR, with a Reinforcement Learning (RL) policy predicated on GPs via their linear combination. The methodology is not only user-friendly but also demonstrates accelerated convergence rates and superior control performance.

\begin{figure*}[!t]
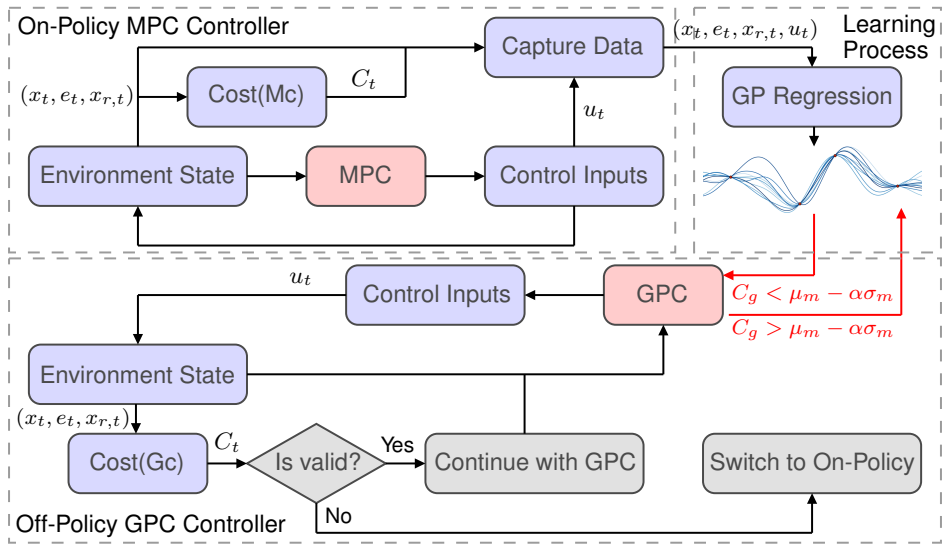

\centering
\include{tikz-helper-updated}
\vspace{-25pt}
\caption{Architecture of the proposed off-policy Gaussian Predictive Control framework, composed of three main components: 1) An On-Policy MPC controller as the foundational layer of the controller; 2) A Learning Process, where the system’s interaction with the environment is observed and are collected and a generalized learning model is trained; 3) An Off-Policy GPC controller, evolved and designed based on Gaussian Processes, that becomes the primary controller once it has adequately learned and evolved from the acquired data.}
\label{fig:architecture}
\vspace{-10pt}
\end{figure*}

Our implementation of Model Predictive Control (MPC) in this work is largely influenced by~\cite{NMPC}, where the authors introduced a model-based non-linear controller designed to track trajectories while concurrently mitigating collisions. Our inclination towards utilizing GPs as a learning method is inspired by~\cite{GP-Tutorial}. In this reference, the authors describe the advantages of employing Gaussian Processes for learning control methodologies, emphasizing the limitations of using parametric models in capturing the complex nature of such control methods.

The work detailed in~\cite{Diff-MPC} introduces the concept of leveraging MPC as a differentiable policy class for reinforcement learning within continuous state and action spaces. The authors employ the KKT conditions of convex approximation to differentiate through MPC, enabling the learning of the cost and dynamics of a controller through end-to-end learning. Their explorations are primarily grounded in imitation learning within pendulum and cart-pole domains, demonstrating that their formulated MPC policies not only exhibit heightened data efficiency compared to generic neural networks but also surpass traditional system identification approaches, especially in scenarios characterized by unreliable experts.

Weidmann et al.~\cite{APG} propose an Analytic Policy Gradient (APG) method tailored for control design in robotic systems. This method is designed to address the delicate balance between accuracy and computational efficiency. By leveraging differentiable simulators, the authors train a controller offline, utilizing gradient descent on the tracking error to attain high accuracy.
Here, APG surpasses both model-based and model-free RL methods in tracking error and mirrors the performance of MPC, albeit with significantly reduced computation time.
Boedecker et al.~\cite{Sparse-GP} introduce a fully automated strategy for optimal control of non-linear systems, incorporating Gaussian Process Regression (GPR) and a modified iterative LQR formulation. This algorithm constructs a non-parametric model of system dynamics and executes receding horizon control, demonstrating data efficiency and the capability to function under real-time constraints.
The incorporation of an exploration strategy founded on GPR variance further elevates the algorithm's performance. This exploration unveils the significant potential of GPR-based methodologies in optimizing control for non-linear systems.

Torrente et al.~\cite{DD-MPC} put forth a groundbreaking method to model the intricate aerodynamic effects impacting quadrotors at elevated speeds, employing Gaussian Processes (GP). Integrating the GP model within a Model Predictive Controller enables the achievement of streamlined and accurate real-time feedback control, subsequently leading to a reduction of up to 70\% in trajectory tracking error.
This study shows the substantial enhancements GP models can render in high-speed trajectory tracking for quadrotors.
Bouffard et al.~\cite{LB-MPC} illustrate a real-time application of learning-based model predictive control (LBMPC) on a quadrotor helicopter. By combining statistical learning with control engineering, LBMPC elevates safety, robustness, and convergence.
Demonstrating resilience to mis-learning, the paper further exemplifies the applicability of LBMPC in a complex robotic task, wherein the quadrotor adeptly catches a ball following an unpredictable trajectory.

Hewing et al.~\cite{Cautious-GPC} introduce an MPC methodology, incorporating a nominal system paired with an additive nonlinear component of dynamics represented as a GP. This application of GP regression models residual uncertainties inherent in the system, thereby facilitating cautious control. The paper develops approximation methodologies for state distribution propagation and provides a well-founded framework for formulating the chance-constrained MPC problem.
This work underscores the advantages of employing GP regression in modeling nonlinear dynamical systems for control purposes.

Thus far, existing studies predominantly focus on learning either the uncertainties inherent in the dynamics of a robot, the environment, or, more broadly, a control method tailored for a specific application. In contrast, our work introduces a novel off-policy method designed to comprehend the behavior of optimal control methods, encompassing both robot and environment dynamics. We put forth a unified architecture devided into on-policy and off-policy controllers, facilitating seamless deployment into real-time systems.

\section{Proposed Framework}\label{sec:method}

Fig.~\ref{fig:architecture} illustrates the overall architecture of the proposed controller, which is divided into three main components. Initially, an On-Policy controller, functioning as an MPC, is set up to control the system along the desired trajectory while avoiding obstacles. This controller serves as the foundational layer, upon which a GP-based off-policy controller is subsequently trained to emulate its behavior. During the system's interaction with the environment, pertinent data -- including system states, environment states, and control law -- are collected. Subsequently, a generalized learning process is established, not incorporating any information on system dynamics, and is trained to imitate the On-Policy controller. The final component is the fully evolved GP-based controller, which becomes the primary control mechanism once sufficient data have been acquired, superseding the on-policy controller.\looseness=-1

The following subsections describe the utilization of model predictive control, followed by a background on Gaussian Processes, and introduce the proposed Gaussian Predictive Controller.

\subsection{Model Predictive Control} \label{sec:mpc}

We employ nonlinear model predictive control (NMPC) as an on-policy controller for a robotic system, drawing inspiration from~\cite{NMPC}. At every time step, we solve the optimal control problem as stated:
\begin{align} \label{eq:OCP}
  \min_{\bm{x,u}} \int_{t=0}^T &{\Big(J_{x}\big(\bm{x}(t),\bm{x_r}(t)\big) + J_c\big(\bm{x}(t)\big)\Big)} \;{dt} \\
  \textrm{subject to:} \quad &\bm{\dot{x}} = f(\bm{x,u}) \nonumber \\
    & \bm{u}(t) \in \mathcal{U} \nonumber \\
    & G(\bm{x}(t)) \leq 0 \nonumber \\
    & \bm{x}(0) = \bm{x}(t_0) \nonumber 
\end{align}
Here, $f$ denotes the state space model of the mobile robot, while $J_x$ and $J_c$ represent the cost functions for trajectory tracking and collision, respectively. $G$ indicates the constraints imposed on the states, and $\mathcal{U}$ represents the permissible range of control inputs. The cost function $J_x$ quantifies the deviation of the predicted state $\bm{x}$ from the desired state $\bm{x_r}$ and is formulated as follows:
\begin{equation} \label{eq:jx}
   J_{x}\big(\bm{x}(t),\bm{x_r}(t)\big) = \| \bm{x}(t)-\bm{x_r}(t)\|^2_{Q_{x}}
\end{equation}
In this equation, $Q_x$ serves as a tuning parameter. Same way, $J_c$ is associated with collision cost and is expressed by:
\begin{equation} \label{eq:jc}
   J_c(\bm{x}(t)) = \sum_{j=1}^{N_{\textrm{obstacle}}} \dfrac{Q_{c,j}}{1+exp\big(k_j (d_j(t)-r_{th,j}(t))\big)}
\end{equation}
Here, $Q_{c,j}$ is a tuning parameter, $d_{j}(t)$ is the Euclidean distance between the robot and the $j^{th}$ obstacle, and $r_{th,j}(t)$ is a threshold distance beyond which collision is imminent. This threshold distance is pivotal, ensuring collision avoidance by maintaining a safe proximity between obstacles and the robot.\looseness=-1

In implementing this controller, accurately predicting the movement of dynamic obstacles is crucial. For simplicity, a constant velocity model can be employed to predict the future trajectory of an obstacle within a predefined time window, as described by equation:
\begin{equation} \label{eq:pred}
   p_j(t) = p_j(t_0)+v_j(t_0)(t-t_0+\delta)
\end{equation}
In this equation, $\delta$ accounts for delays induced by computation and communication processes. To optimize performance, the variable $t$ can be assigned a constant value, representing a fixed window of time.

The implementation of the above controller starts with the discretization of the integral in~\eqref{eq:OCP}, leading to the following optimization:
\begin{align}\label{eq:opt}
J(\bm{u}) = &\sum_{t=0}^{T} C(\bm{x}_t, \bm{x}_{r,t},\bm{e}_t) \\
\textrm{subject to:} \quad & \bm{x}_{t+1} = F(\bm{x}_t,\bm{u}_t) \nonumber \\
    & \bm{u}(t) \in \mathcal{U} \nonumber \\
    & \bm{x}(0) = \bm{x}(t_0) \nonumber 
\end{align}
Here, $\mathbf{x}_t$ denotes the state vector of the robot at time step $t$, $\mathbf{x}_{r,t}$ represents the reference trajectory of the robot, $\mathbf{e}_t$ denotes the state vector of the environment, and $\mathbf{u}$ represents the control inputs.

The involved cost function in~\eqref{eq:opt} may be non-convex due to which the Sequential Least-Squares Quadratic Programming (SLSQP) method~\cite{kraft1988software} is employed for minimizing it. SLSQP is a versatile iterative optimization algorithm designed to find the variable values that minimize the objective function while concurrently satisfying a set of equality and inequality constraints.
At each iteration, the objective function and constraints are approximated using quadratic models around the current point, solving the quadratic sub-problem to determine the step minimizing the model under the constraints. The optimized variables are then updated with this computed step, taken in a direction that reduces the objective function's value, and the process iterates until convergence, ensuring training proceeds at feasible speeds.

\subsection{Gaussian Processes}\label{sec:gp}

In the proposed framework, Gaussian Processes (GPs) play a pivotal role in forging a learning-based approach to understand and replicate the behavior of Model Predictive Control (MPC). Specifically, GPs are deployed to model the optimization dynamics inherent in optimal control strategies such as MPCs. This is facilitated by leveraging Gaussian Process Regression to understand the knowledge from the MPC algorithm's interaction with a given robotic system, allowing for an accurate learning of the system’s dynamics and its interaction with the environment.

A GP is a flexible and non-parametric statistical model that defines a distribution over functions. Its use and properties for modeling are reviewed in~\cite{pred_gp,GPML}. A GP is defined by a mean function and a covariance (kernel) function. The mean function represents the expected value of the function at each input point, while the covariance function quantifies the relationships between data points, capturing smoothness and correlations. It possesses the property that any subset of states is represented as a collection of random variables, all of which follow a joint multivariate Gaussian distribution. For the collection of $d$ random variables  $\bm{X_i},\dots,\bm{X_j} \sim \mathbb{N}(\bm{\mu},\Sigma)$, with mean $\bm{\mu} \in \mathbb{R}^d$ and covariance $\Sigma \in \mathbb{R}^{d \times d}$, joint Gaussian density is represented using:
\begin{equation} \label{eq:gd}
   p(\bm{x}) = \frac{1}{\sqrt{(2\pi)^d\det|\Sigma|}} \exp\biggl(-\frac{1}{2}(\bm{x}-\bm{\mu})^T\Sigma^{-1}(\bm{x}-\bm{\mu})\biggr)
\end{equation}

Given a GP, specified by mean ${m}(\cdot)$ and covariance $k(\cdot,\cdot)$, a function $f(\bm{x})$ can be sampled at any point $\bm{x}\in\mathbb{R}^d$ from the GP, according to \eqref{eq:gpr}, where ${m}(\bm{x}) = \mathbb{E}[f(\bm{x})]$ represents the expected value of $f(x)$, and $k(\bm{x},\bm{x'}) = \mathbb{E}[(f(\bm{x})-{m}(\bm{x}))(f(\bm{x'})-{m}(\bm{x'}))]$. Here, $\mathbb{E}$ denotes the expected value. 
\begin{equation} \label{eq:gpr}
  f(\bm{x}) \sim \mathcal{GP}\big({m}(\bm{x}),k(\bm{x},\bm{x'})\big)
\end{equation}
The formulation of Gaussian Processes outlined above facilitates leveraging all available data points when predicting the output at any new state, employing it as a conditional probability via Gaussian Process Regression (GPR). GPR stands as a powerful non-parametric learning technique capable of capturing complex relationships within the data without making strong assumptions about the underlying model. The central concept of GPR is to utilize the predictive distribution of the GP to estimate the target variable for a given input feature set. Moreover, GPR provides not only point estimates but also quantifications of predictive uncertainties. This dual capability renders it invaluable for decision-making processes in uncertain environments. The following section describes the deployment of GPR as a controller in this work.

\subsection{Off-Policy Gaussian Predictive Control}\label{sec:gpc}

The off-policy Gaussian Predictive Control developed in this work is formulated by the fusion of learning-based techniques with MPC, in which Gaussian Processes (GPs) are employed to model and learn the optimal control strategies inherent in MPC. The controller is trained off-policy, utilizing the data collected from the on-policy MPC interactions with the environment, including system and environment states and control law, without incorporating the system dynamics explicitly, making it broadly applicable to a myriad of robotic platforms. This approach allows the controller to learn and adapt optimally to diverse trajectories and environments, eventually superseding the on-policy controller upon acquiring sufficient data. The integration of GPs not only enables accurate trajectory tracking but also offers enhanced performance in real-time applications across various robotic systems.\looseness=-1

In contrast to previous works where Gaussian Processes (GPs) have primarily been utilized to deduce the dynamics function $F(\mathbf{x}, \mathbf{u})$ or to predict the discrepancies between the dynamic model and reality, in this work, the GP has been strategically employed to approximate the control policy. This is represented as:
\begin{equation}
\min_{\bm{u}} J(\bm{u}|\bm{x}_{t},\bm{x}_{r,t},\bm{e}_t) \approx \mathcal{G}P\big({m}(\bm{x}),k(\bm{x},\bm{x'})\big)
\end{equation}
In this equation, the minimization with respect to $\bm{u}$ of $J(\bm{u}|\bm{x}_{t},\bm{x}_{r,t})$ is the output of the cost minimization operation, while $\bm{x},\bm{x}_r,\bm{e}_t$ represent the state of the controller, the reference trajectory, and the state of the environment, respectively.\looseness=-1

To avoid incorporating any model into the regression, a zero-mean Gaussian Process is employed, meaning $\bm{\mu}(\bm{x}_{t+1})=0$. Consequently, for all Gaussian priors, ${m}(\bm{x})=0$, ensuring that the process remains uninfluenced by any presumptive modeling, and allowing for a more unbiased and untainted approximation.
Thus, the next state (at $t+1$) is predicted solely based on prior inputs to Gaussian Process Regression.

To utilize GP as a controller, it is trained using features such as $\bm{x}_t,\,\bm{e}_t,\,\bm{x}_{r,t}$ and corresponding labels $\bm{u}_t$, which is the control input. With GP configured to approximate the model predictive controller, it becomes feasible to employ it as a real-time controller. This is because the GP can generate control inputs at a much faster rate, enhancing the overall efficiency and responsiveness of the system.

\subsection{Switching from MPC to GPC}\label{sec:switch}

To facilitate a seamless transition from Model Predictive Control (MPC) to the Gaussian Process-based controller (GPC), a criterion is introduced to dynamically select the most suitable controller based on performance metrics. The selection criterion is determined by evaluating the cost associated with each controller. Specifically, if the cost associated with the GPC controller, denoted as $C_{g}$, falls below a certain threshold relative to the mean ($\mu_m$) and standard deviation ($\sigma_m$) of the costs incurred by the MPC (that is used to train the model), the system transitions to GPC control. The cost $C_{g}$ is computed the same as the MPC cost in~(\ref{eq:OCP}), and is given by:
\begin{equation}\label{eq:cg}
    C_g = J_{x}\big(\bm{x}(t),\bm{x_r}(t)\big) + J_c\big(\bm{x}(t)\big)
\end{equation}
The switching criterion is defined as:
\begin{equation}\label{eq:switch_condition}
C_g <  \mu_m - \alpha\sigma_m
\end{equation}

Here, $\alpha$ serves as a tunable parameter, allowing flexibility in adjusting the sensitivity of the switching decision. The rationale behind this criterion is to ensure a smooth and effective transition, enabling the GPC controller to take over when its performance consistently surpasses that of the MPC controller, as indicated by the cost comparison.

In summary, the proposed research introduces a novel approach to robot control using GP regression. By combining on-policy and off-policy data and leveraging the expressive power of Gaussian Processes, the controller can effectively adapt to the complexities of different environments, leading to enhanced robotic performance. This strategy ensures a seamless and real-time generation of control inputs, allowing the GP to act as a real-time controller and make efficient, informed decisions, effectively navigating the robot through unpredictable terrains and environments.

\section{Experiment Setup} \label{sec:env}

The proposed control algorithm is evaluated using a simulated differential drive mobile robot (DDMR). Initially, the robot utilizes Model Predictive Control (MPC) to track various trajectories, while also avoiding a moving obstacle, which is programmed to follow different paths. During these operations, all essential data
are collected and stored for subsequent Gaussian Process Regression (GPR). The stored input data encompasses the robot state, trajectory, and obstacle state, and the corresponding labels are the control laws applied.
In the final testing phase, Gaussian Predictive Control (GPC) is employed in place of MPC to track unseen trajectories. The performance of GPC is then compared to that of MPC in terms of accuracy in trajectory tracking and computation time to assess the effectiveness of the proposed approach in learning and emulating the behavior of the initial on-policy controller and optimizing the performance metrics.

The dynamic model employed for DDMR in this study is derived from the models presented in~\cite{dynamics1} and~\cite{dynamics2}. This dynamic model is represented in the manipulator form as shown in \eqref{eq:manipulator_form}. The control inputs for this robot model consist of the torques applied to the left and right wheels, denoted by $\bm{T}$.
\begin{equation}\label{eq:manipulator_form}
M(\bm{q})\Ddot{\bm{q}} + B(\bm{q},\dot{\bm{q}})-C^T(\bm{q})\bm{\lambda} = \bm{T}
\end{equation}
In this equation, $\bm{q}$ represents the set of generalized coordinates and is defined as 
 $\bm{q}=
\begin{bmatrix}
    x &
    y &
    \theta &
    \phi_1 &
    \phi_2 
\end{bmatrix}^T$
Here, \( x, y, \) and \( \theta \) correspond to the position and orientation of the DDMR, while \( \phi_1 \) and \( \phi_2 \) represent the angular positions of the right and left wheels, respectively.

Moreover, in \eqref{eq:manipulator_form}, $M(\bm{q})$ represents the symmetric positive definite mass matrix, and $B(\bm{q,\dot{q})}$ denotes the Coriolis matrix. Given that the subject is a mobile robot, there is no gravity vector included. Both the mass and Coriolis matrices can be expressed as follows:
\[
\setlength\arraycolsep{1.25pt}
M(\bm{q}){=}
\begin{bmatrix}
    m_T & 0 & {-}m_Bd\sin{(\theta)} & 0 & 0\\
    0 & m_T & m_Bd\cos{(\theta)} & 0 & 0\\
    {-}m_Bd\sin{(\theta)} & m_Bd\cos{(\theta)} & I_T & 0 & 0\\
    0 & 0 & 0 & I_{yy}^B & 0\\
    0 & 0 & 0 & 0 & I_{yy}^B
\end{bmatrix}
\]
\[
B(\bm{q},\dot{\bm{q}}) = -m_Bd\dot{\theta}^2
\begin{bmatrix}
    cos(\theta)\\
    sin(\theta)\\
    0\\
    0\\
    0 
\end{bmatrix}
\]
where $m_B$, $m_w$, and $m_T$ denote the mass of the chassis, wheel, and the total mass of the DDMR, respectively; $I^B$ and $I^T$ represent the moment of inertia of the chassis and the total moment of inertia of the DDMR, respectively; and
$d$ is the distance from the center of each wheel to the center of mass of the chassis.

Additionally, $\bm{\lambda}$ is denoted as the vector of constrained forces, and $C(\bm{q})$ is the matrix associated with constraints in place. The constraints are presumed to be formulated as $ C(\bm{q})\bm{\dot{q}} = 0$.
For the DDMR, both $C(\bm{q})$ and $\bm{\lambda}$ have specific expressions that consider the unique structure and operational constraints of this type of robot:
\[
C(\bm{q}) =
\begin{bmatrix}
    cos(\theta) & sin(\theta) & 0 & \frac{\rho}{2} & -\frac{\rho}{2}\\
    -sin(\theta) & cos(\theta) & 0 & 0 & 0\\
    0 & 0 & 1 & \frac{\rho}{2W} & \frac{\rho}{2W}
\end{bmatrix}
\]
and
\begin{multline}
    \bm{\lambda}{=}{-}[C(\bm{q})M^{-1}(\bm{q})C^T(\bm{q})]^{-1} \\ [C(\bm{q})M^{-1}(\bm{q})(\bm{T}-B(\bm{q},\dot{\bm{q}})){+} \dot{C}(\bm{q})\dot{\bm{q}}]
\end{multline}

\begin{figure}[!htbp]
\centerline{\includegraphics[width=\columnwidth,trim={20pt 20pt 20pt 25pt},clip]{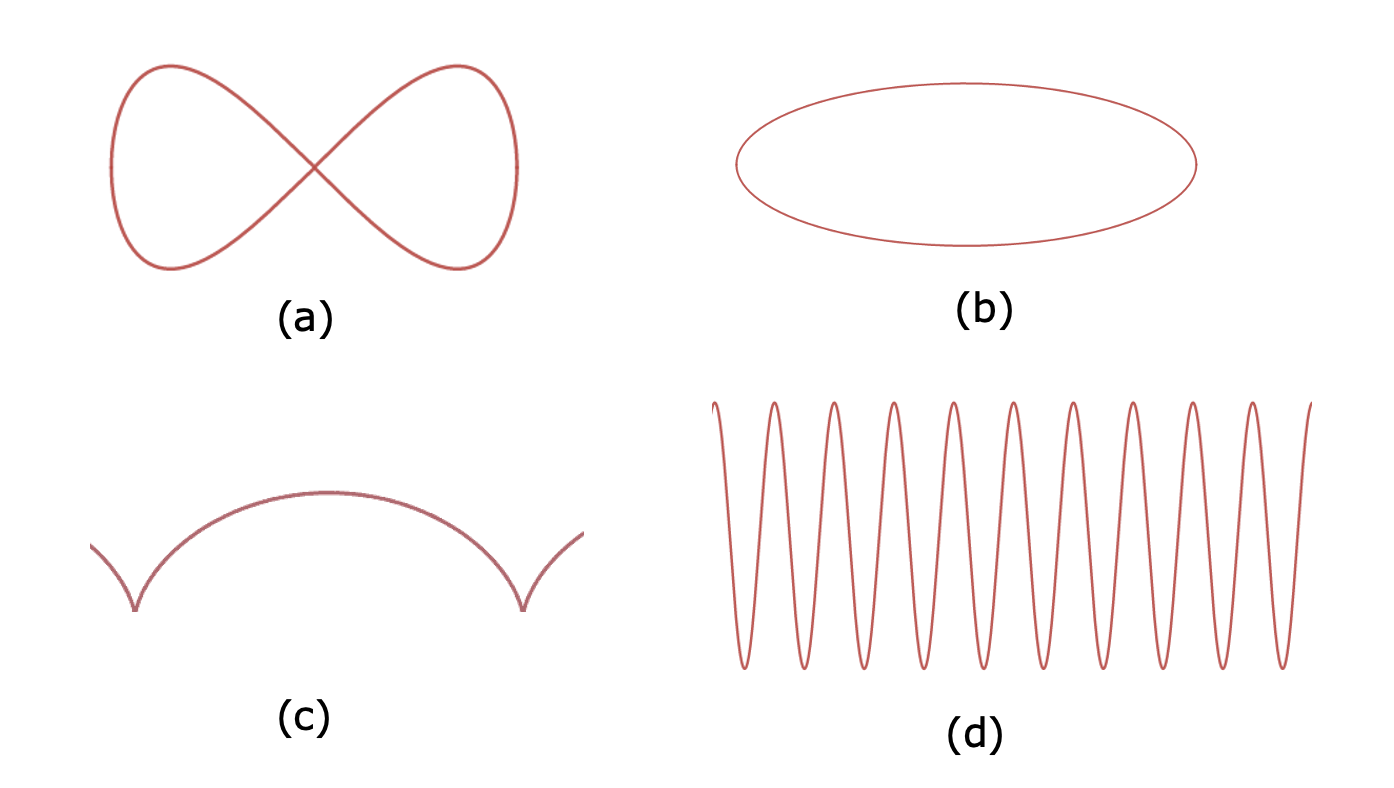}}
\vspace{-5pt}
\caption{Plot of trajectories: (a) Leminscate of Gerono;
(b) Ellipse; (c) Cycloid; and
(d) Sine wave.}
\label{fig:fuzzy-anfis}
\vspace{-10pt}
\end{figure}

\begin{figure}[!htbp]
\centerline{\includegraphics[width=\columnwidth,trim={10pt 20pt 20pt 20pt},clip]{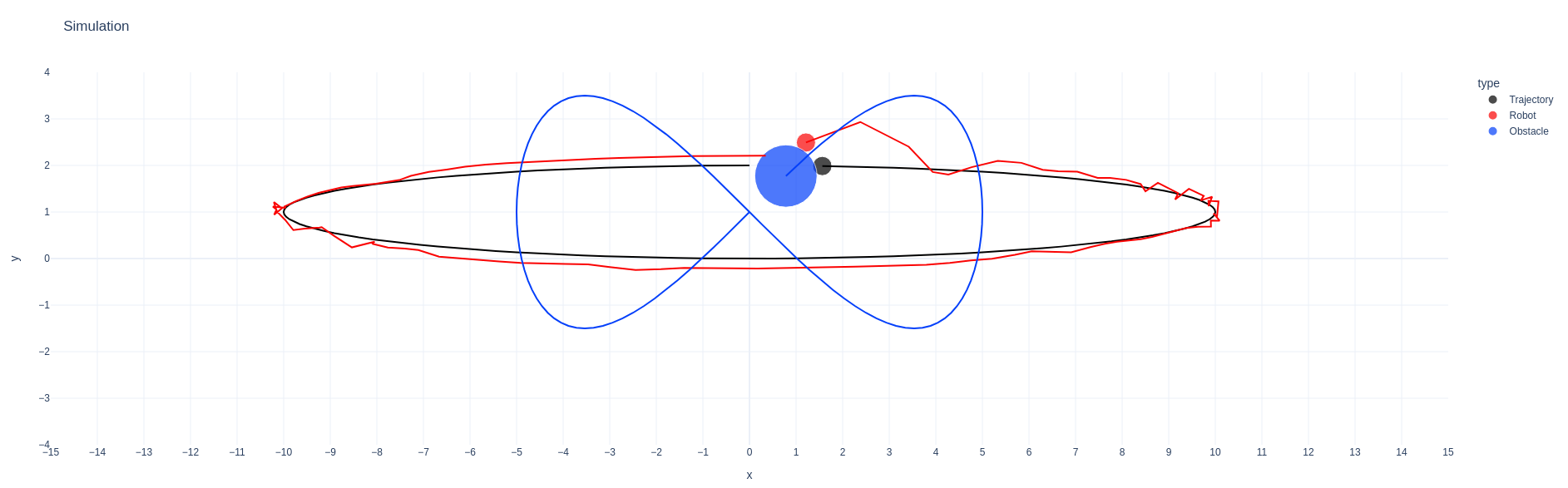}}
\vspace{-5pt}
\caption{MPC performance on the robot to follow an elliptical reference trajectory while avoiding an obstacle trajectory that is a Leminscate of Geron.}
\label{fig:env_5}
\vspace{-10pt}
\end{figure}

\begin{figure}[!htbp]
\centerline{\includegraphics[width=\columnwidth,trim={10pt 20pt 20pt 20pt},clip]{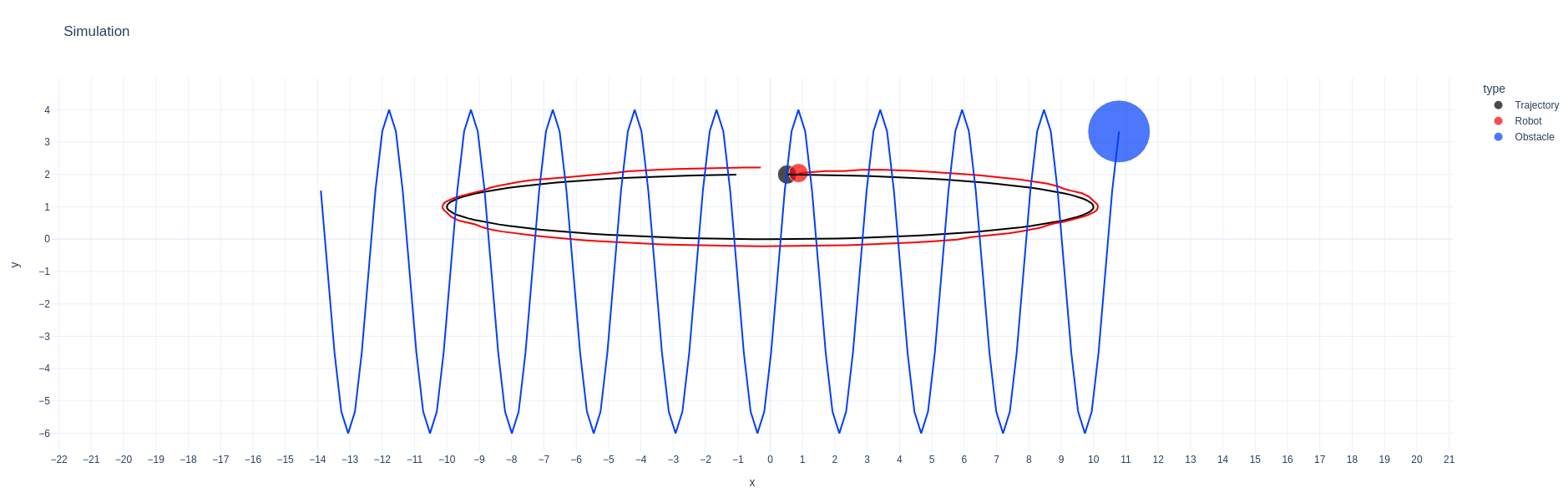}}
\vspace{-5pt}
\caption{MPC performance on the robot to follow a sine wave reference trajectory while avoiding an elliptical obstacle trajectory.}
\label{fig:env_2}
\end{figure}

In order to conduct simulations and collect the necessary data, 10 diverse trajectories were generated, deriving from four foundational curves: sine, Lemniscate of Gerono, ellipse, and cycloid, as depicted in Fig. \ref{fig:fuzzy-anfis}. The mathematical representations for these curves are provided by equations~\eqref{eq:c1}, ~\eqref{eq:c2},~\eqref{eq:c3}, and~\eqref{eq:c4}. The testing environments, designed to simulate two distinct scenarios, with MPC's performance, are illustrated in Figs.~\ref{fig:env_5} and~\ref{fig:env_2}.
\begin{equation}\label{eq:c1}
   x^4-x^2+y^2=0
\end{equation}
\begin{equation}\label{eq:c2}
   \frac{x^2}{a}+\frac{y^2}{b}= 1
\end{equation}
\begin{equation}\label{eq:c3}
   y = \sin{x}
\end{equation}
\begin{equation}\label{eq:c4}
   x = r\cos^{-1}({1-\frac{y}{r}}) - \sqrt{y(2r-y)}
\end{equation}

\vspace{-10pt}
\vspace{-10pt}
\vspace{-10pt}

For each environment, we select two of the 10 generated curves, allocating one for the robot trajectory and the other for the obstacle path, culminating in a total of 90 distinct environments. We documented the behaviors of MPC within all these scenarios, recording them along with the states of the environments. The newly devised off-policy controller was subsequently trained utilizing Gaussian Process Regression, incorporating a Radial Basis Function (RBF) kernel, represented by \eqref{eq:rbf}.
\begin{equation} \label{eq:rbf}
   k_{RBF}(\bm{x},\bm{x}') = \exp\biggl(-\frac{1}{2}(\bm{x}-\bm{x'})\Phi^{-2}(\bm{x}-\bm{x'})\biggr)
\end{equation}
In this equation, $\Phi$ is a diagonal matrix of length scales, with $\Phi \in \mathbb{R}^{d\times d}$.

\section{Results} \label{sec:results}

In this section, we present the results from extensive simulation experiments to evaluate the performance of the proposed Gaussian Process Control (GPC) against Model Predictive Control (MPC). These experiments covered various environments and provided insights into different aspects of the GPC controller.

\begin{figure}[!htbp]
\centerline{\includegraphics[width=\columnwidth,trim={10pt 4pt 10pt 10pt},clip]{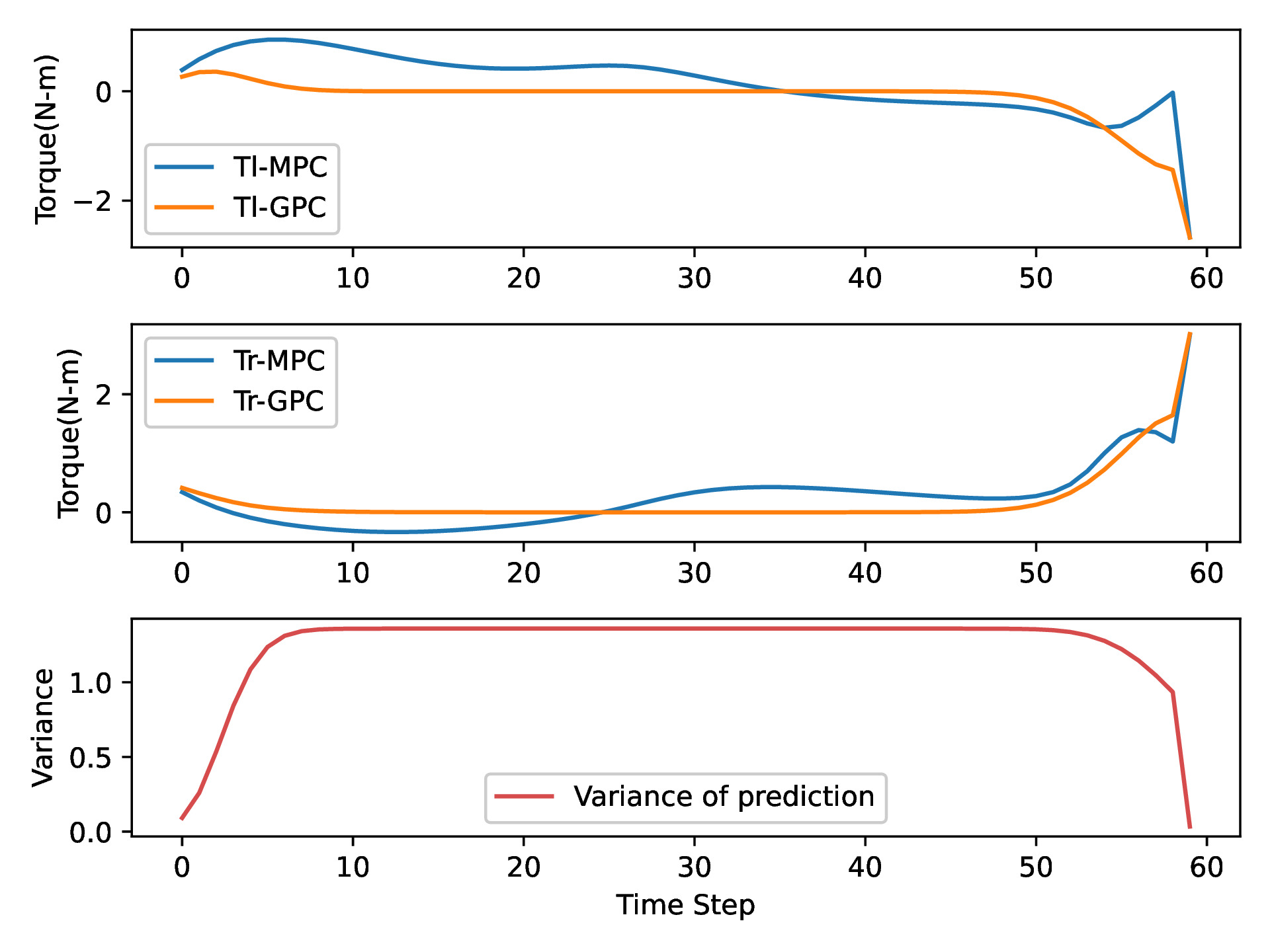}}
\vspace{-10pt}
\caption{The controller was trained on the first half of the environment and tested on the second half. The plots depict the torque commands of the left and right wheels for both MPC and GPC controllers, and the variance computed by the GP regression.}
\label{fig:s1}
\vspace{-10pt}
\end{figure}

Our primary objective was to evaluate the adaptability and learning speed of GPC within its environment. To do this, the controller was trained on a limited portion of an environment, and then the similarity between the MPC and GPC controller actions was compared for the unseen portion of the environment. As can be seen in Fig. \ref{fig:s1}, with training only in the first half of an environment, the control input generated by the GPC was already notably similar to that of the MPC. This testing compared the control inputs by the GPC, given the environment and robot state, with those by the MPC. The figure includes three plots; the first showing the torque command of the left wheel, the second the torque command of the right wheel, and the third depicting the variance computed by the GP regression, indicating the confidence in the output. These findings suggest that GPC is proficient at rapidly learning MPC's control law. However, the variance noted during testing iterations is above 1, indicating that the model may still need to fully grasp the complexities of the MPC's control strategy.

\begin{figure}[!htbp]
\centerline{\includegraphics[width=\columnwidth,trim={10pt 4pt 10pt 10pt},clip]{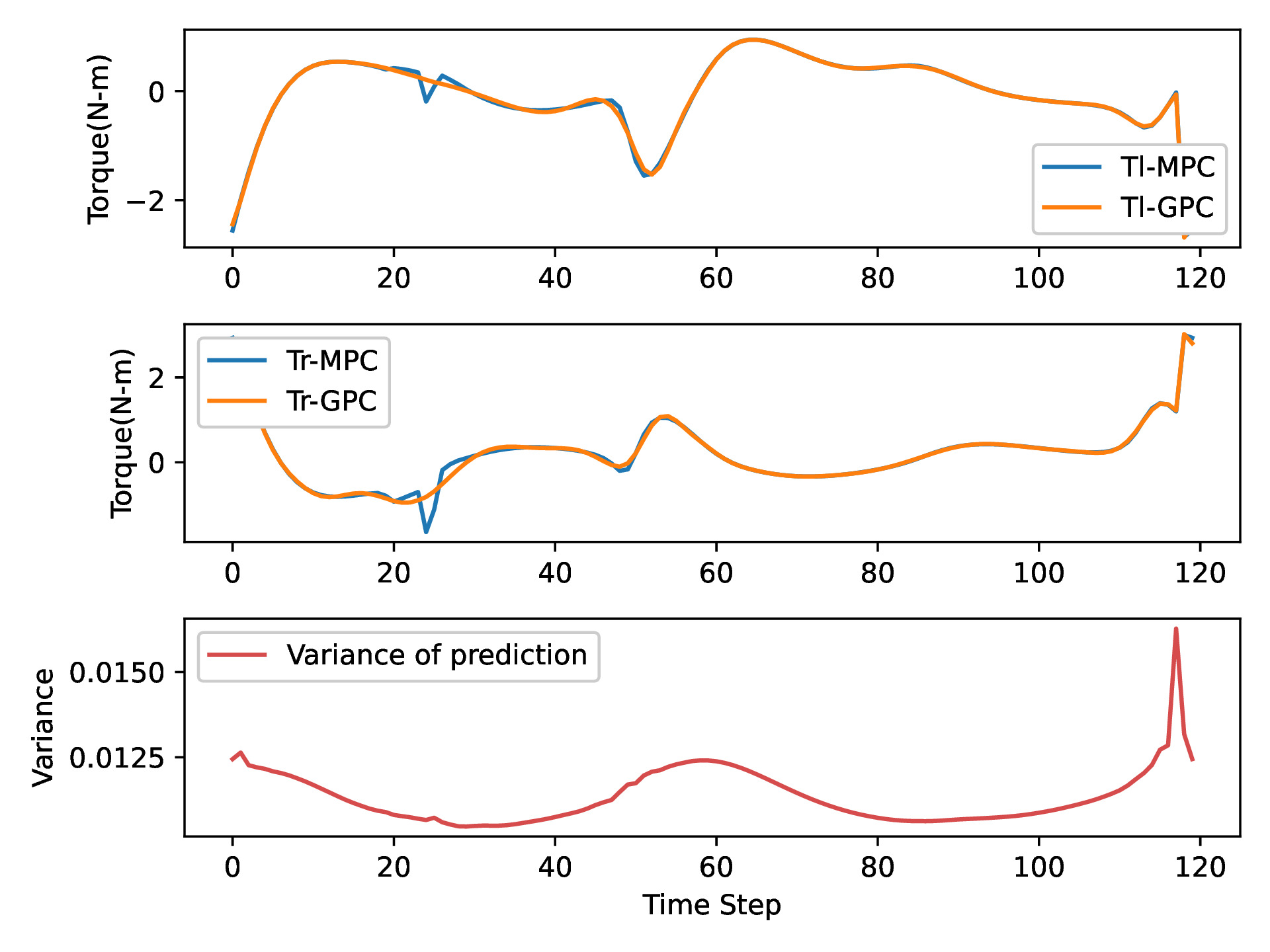}}
\vspace{-10pt}
\caption{The controller was trained in one environment and tested in another environment with slightly modified parameters. The plots depict the torque commands of the left and right wheels for both MPC and GPC controllers, and the variance computed by the GP regression.}
\label{fig:s2}
\vspace{-10pt}
\end{figure}

The subsequent series of experiments focused on evaluating the adaptability of the GPC controller to minor variations in the environment. Fig. \ref{fig:s2} illustrates the performance of the model when trained in one environment and then evaluated in a slightly varied setting. Remarkably, the GPC controller demonstrates a substantial alignment with the MPC’s control strategy, even after exposure to a single environment. This alignment is more noticeable as the model undergoes further training with additional data and is evaluated in a known environment. This highlights the exceptional ability of the GPC to learn MPC's interaction in diverse settings. The exceedingly low variance in GPC’s control inputs indicates its high level of confidence in the derived control law.


\begin{figure}[!htbp]
\centerline{\includegraphics[width=\columnwidth,trim={10pt 4pt 10pt 10pt},clip]{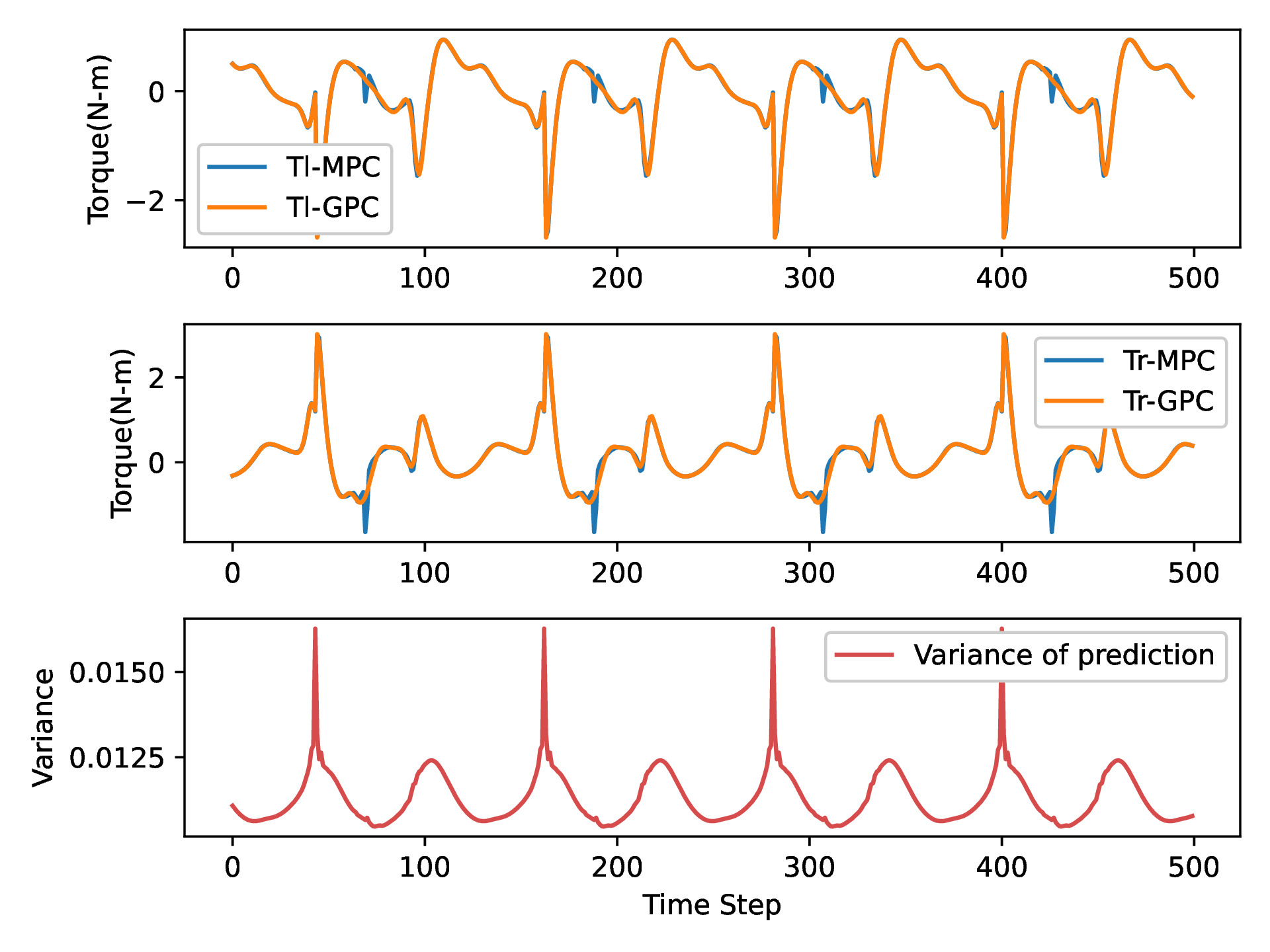}}
\vspace{-10pt}
\caption{The controller was trained across six different environments and subsequently tested in four completely distinct, untrained environments. The plots depict the torque commands of the left and right wheels for both MPC and GPC controllers, and the variance computed by the GP regression.}
\label{fig:s5}
\vspace{-10pt}
\end{figure}

\begin{figure}[!htbp]
\centerline{\includegraphics[width=\columnwidth,trim={10pt 20pt 20pt 20pt},clip]{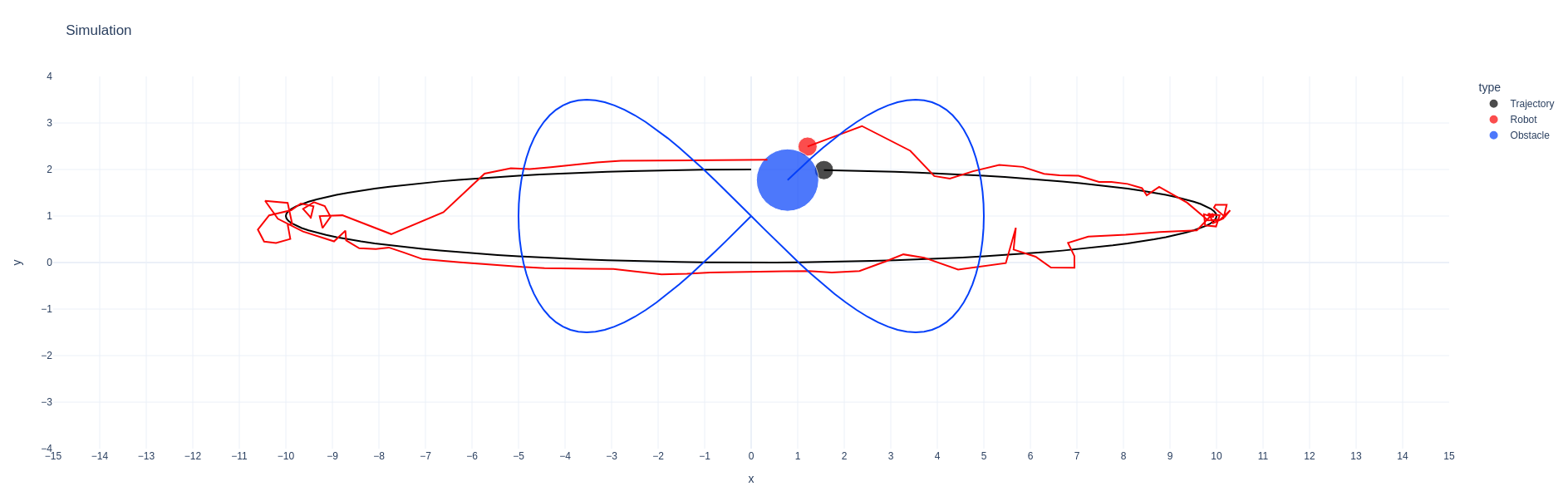}}
\vspace{-5pt}
\caption{GPC performance on the robot in an environment same as Fig.\ref{fig:env_5}.}
\label{fig:gpc_1}
\vspace{-10pt}
\end{figure}

\begin{figure}[!htbp]
\centerline{\includegraphics[width=\columnwidth,trim={10pt 20pt 20pt 20pt},clip]{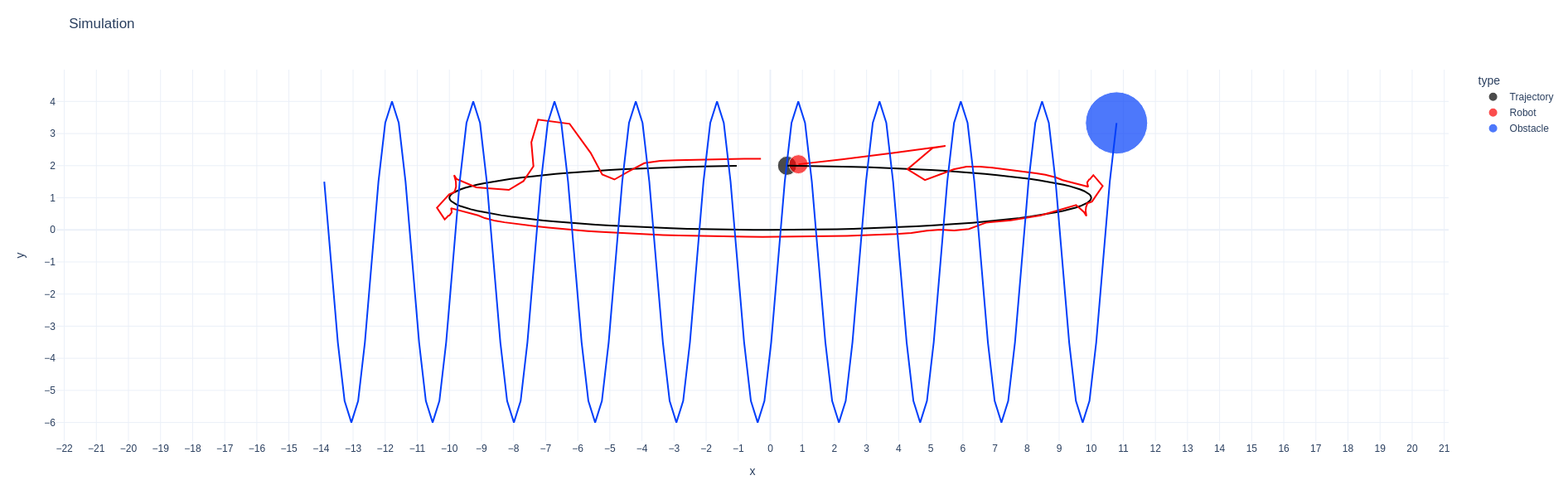}}
\vspace{-5pt}
\caption{GPC performance on the robot in an environment same as Fig.\ref{fig:env_2}.}
\label{fig:gpc_2}
\vspace{-10pt}
\end{figure}

An integral component of our evaluation focused on investigating the model’s ability to generalize. We aimed to determine whether GPC could reliably replicate MPC’s performance in environments it had not been exposed to during training. Fig. \ref{fig:s5} presents the results in untrained, novel environments, demonstrating GPC’s resilient and adaptive performance. The total trajectory tracking cost in six diverse environments for both MPC and GPC is documented in Table \ref{tab:tab3}. It is observable that the trajectory following costs using control inputs from both GPC and MPC are comparable, indicating a close alignment in the control inputs’ impact on the robot state evolution between the GPC and the MPC. This exemplifies the robustness and adaptability of GPC in navigating unknown environments, replicating the directionality and state evolution induced by MPC with notable accuracy.
The trajectory tracking performance of the GPC is presented in Figs. \ref{fig:gpc_1} and \ref{fig:gpc_2}. The results shown in these figures were obtained in the same environments as those in Figs. \ref{fig:env_5} and \ref{fig:env_2}, respectively. It is observed that the trajectory tracked by the GPC closely matches that of the MPC. This suggests that the proposed GPC can serve as an effective substitute for the MPC in trajectory tracking tasks, offering comparable performance without the computational overhead associated with MPC optimization and thus achieving real-time performance.
 
\begin{table}[!t]
\centering
\caption {Comparative analysis of costs incurred by the two controllers in different environments.}
\vspace{-5pt}
\begin{tabular}{|l|c|c|} 
\hline
\textbf{} & \multicolumn{1}{l|}{\textbf{Total cost by MPC}} & \multicolumn{1}{l|}{\textbf{Total cost by GPC}} \\ 
\hline
\textbf{Env1} & 106.1 & 109.7 \\
\hline
\textbf{Env2} & 196.9 & 204.2 \\
\hline
\textbf{Env3} & 2059.2 & 2142.5 \\
\hline
\textbf{Env4} & 1024.5 & 1068.8 \\
\hline
\textbf{Env5} & 1301.4 & 1356.1 \\
\hline 
\textbf{Env6} & 3772.5 & 3925.4 \\
\hline
\end{tabular}
\label{tab:tab3}
\vspace{-5pt}
\end{table}

\begin{table}[!t]
\centering
\caption {Comparative overview of mean and standard deviation for both controllers.}
\vspace{-5pt}
\begin{tabular}{|l|r|r|} 
\hline
\textbf{Property} & \multicolumn{1}{l|}{\textbf{MPC}} & \multicolumn{1}{l|}{\textbf{GPC}}  \\ 
\hline
\textbf{Mean} (in Seconds) & 65.8 & 30.13 \\
\hline
\textbf{Standard Deviation} & 203.98 & 0.0094 \\
\hline
\end{tabular}
\label{tab:tab2}
\vspace{-5pt}
\end{table}
\begin{table}[!t]
\centering
\caption{Comparative analysis of computation times for GPC and MPC across all scenarios.}
\vspace{-5pt}
\begin{tabular}{|l|r|r|} 
\hline
\textbf{Time range(in Seconds)} & \multicolumn{1}{l|}{\textbf{MPC}} & \multicolumn{1}{l|}{\textbf{GPC}}  \\ 
\hline
30-40                           & 77                                & 90                                 \\ 
\hline
40-50                           & 7                                 & 0                                  \\ 
\hline
60-70                           & 2                                 & 0                                  \\ 
\hline
80-90                           & 1                                 & 0                                  \\ 
\hline
510-520                         & 1                                 & 0                                  \\ 
\hline
1920-1930                       & 1                                 & 0                                  \\
\hline
\end{tabular}
\label{tab:tab1}
\vspace{-10pt}
\end{table}

Finally, a comparative evaluation of the computational efficiency of both GPC and MPC controllers was conducted across all tested environments. Tables \ref{tab:tab2} and \ref{tab:tab1} provide a comparative overview, highlighting the distribution of computation times for both controllers. Notably, MPC demonstrates substantially elevated average computation times, accompanied by a significant increase in variance. This variability implies that MPC’s computation time may vary widely, contingent on the specific environment it interacts with. Conversely, GPC stands out by not only delivering fast control input computations but also by sustaining nearly uniform execution times across the board. This consistency positions GPC as a preferable alternative for real-time control applications, promising steady and efficient performance across diverse environments.

\section{Conclusions}
\label{sec:conclusions}
In conclusion, this study introduced an off-policy Gaussian Predictive Control (GPC) methodology that successfully learns optimal control strategies and operates independently of inherent robot dynamics. The research dissected this methodology into three core components: learning optimal control strategies, adapting to different environments, and ensuring computational efficiency. Through extensive simulations on a differential drive robot, we demonstrated the model’s adaptability and computational efficiency, showcasing its superiority or parity with Model Predictive Control (MPC) in tracking accuracy and obstacle avoidance. This methodology, with its consistent compute time and resilience to variations in the training environment, emerges as a promising and robust solution for a variety of complex applications, especially in real-time, safety-critical scenarios. The approach balances computational efficiency with adaptability, extending its applicability to diverse robotic platforms and environments.

\bibliographystyle{IEEEtran}
\bibliography{bibtex}

\end{document}